\documentclass[journal]{IEEEtran}
\usepackage[utf8]{inputenc}
\usepackage{multirow}
\usepackage{diagbox}
\usepackage[numbers, compress]{natbib}

\usepackage{amssymb}
\usepackage{booktabs,caption}
\usepackage{graphicx}
\usepackage{color}
\usepackage[english]{babel}
\usepackage{xspace}
\usepackage{subfig}
\usepackage{amsmath}

\makeatletter
\DeclareRobustCommand\onedot{\futurelet\@let@token\@onedot}
\def\@onedot{\ifx\@let@token.\else.\null\fi\xspace}
\def\eg{\emph{e.g}\onedot} 
\def\ie{\emph{i.e}\onedot}

\makeatother

\begin{document}
\title{Self-supervised Feature Learning via Exploiting Multi-modal Data for Retinal Disease Diagnosis}

\author{Xiaomeng Li, Mengyu Jia, Md Tauhidul Islam, Lequan Yu,
	and Lei Xing
	
\thanks{Copyright (c) 2019 IEEE. Personal use of this material is permitted. However, permission to use this material for any other purposes must be obtained from the IEEE by sending a request to pubs-permissions@ieee.org.

This work was partially supported by a Faculty Research Award from Google Inc.
	
X. Li, M. Jia, M. Islam, L. Yu, and L. Xing are with the Department of Radiation Oncology, Stanford University, United States (e-mail: {xmengli, jeremy18, tauhid, lequany, lei}@stanford.edu). Corresponding author: Lei Xing.
}}


%

\markboth{}%
{Shell \MakeLowercase{\textit{et al.}}: Bare Demo of IEEEtran.cls for IEEE Journals}
%



\maketitle

\newcommand{\para}[1]{\vspace{.05in}\noindent\textbf{#1}}
\newcommand{\revise}[1]{{\color{black}{#1}}}
\newcommand{\ylq}[1]{{\color{red}{LQ:#1}}}

\begin{abstract}	
	The automatic diagnosis of various retinal diseases from fundus images is important to support clinical decision-making.
	However, developing such automatic solutions is challenging due to the requirement of a large amount of human-annotated data.
	Recently, unsupervised/self-supervised feature
	learning techniques receive a lot of
	attention, as they do not need massive annotations. 
	Most of the current self-supervised methods are analyzed with single imaging modality and there is no method currently utilize multi-modal images for better results. 
	Considering that the diagnostics of various vitreoretinal diseases can greatly benefit from another imaging modality, \eg, FFA, this paper presents a novel self-supervised feature learning method by effectively exploiting multi-modal data for retinal disease diagnosis.
	To achieve this, we first synthesize the corresponding FFA modality and then formulate a patient feature-based softmax embedding objective.
	Our objective learns both modality-invariant features and patient-similarity  features. 
	Through this mechanism, the neural network captures the semantically shared information across different modalities and the apparent visual similarity between patients. 
	We evaluate our method on two public benchmark datasets for retinal disease diagnosis.
	The experimental results demonstrate that our method clearly outperforms other self-supervised feature learning methods and is comparable to the supervised baseline.
	Our code is available at GitHub\footnote{Our code is available at https://github.com/xmengli999/self\_supervised}.
\end{abstract}


\section{Introduction}

\IEEEPARstart{C}{olor} fundus photography has been widely used in clinical practice to evaluate various conventional ophthalmic diseases,~\eg, age-related macular degeneration (AMD)~\cite{age2000risk}, pathologic myopia (PM)~\cite{morgan2012myopia}, and diabetic retinopathy~\cite{zhou2018multi,li2019canet}. 
Recently, deep learning has shown very good performance on a variety of automatic ophthalmic disease detection problems from fundus images~\cite{sakaguchi2019fundus,peng2019deepseenet,virmani2019pnn}, and these techniques can help ophthalmologists in decision making.
The success is attributed to the learned representative features from fundus images, which requires a large amount of training data with massive human annotations.
However, it is tedious and expensive to annotate the fundus images, since experts are needed to provide reliable labels.
Hence, in this paper, our goal is to learn the representative features from data itself, without any human annotations. 
Then, the learned representations are evaluated on the fundus image classification tasks. 

\begin{figure}[t]
	\centering
	\includegraphics[width=0.5\textwidth]{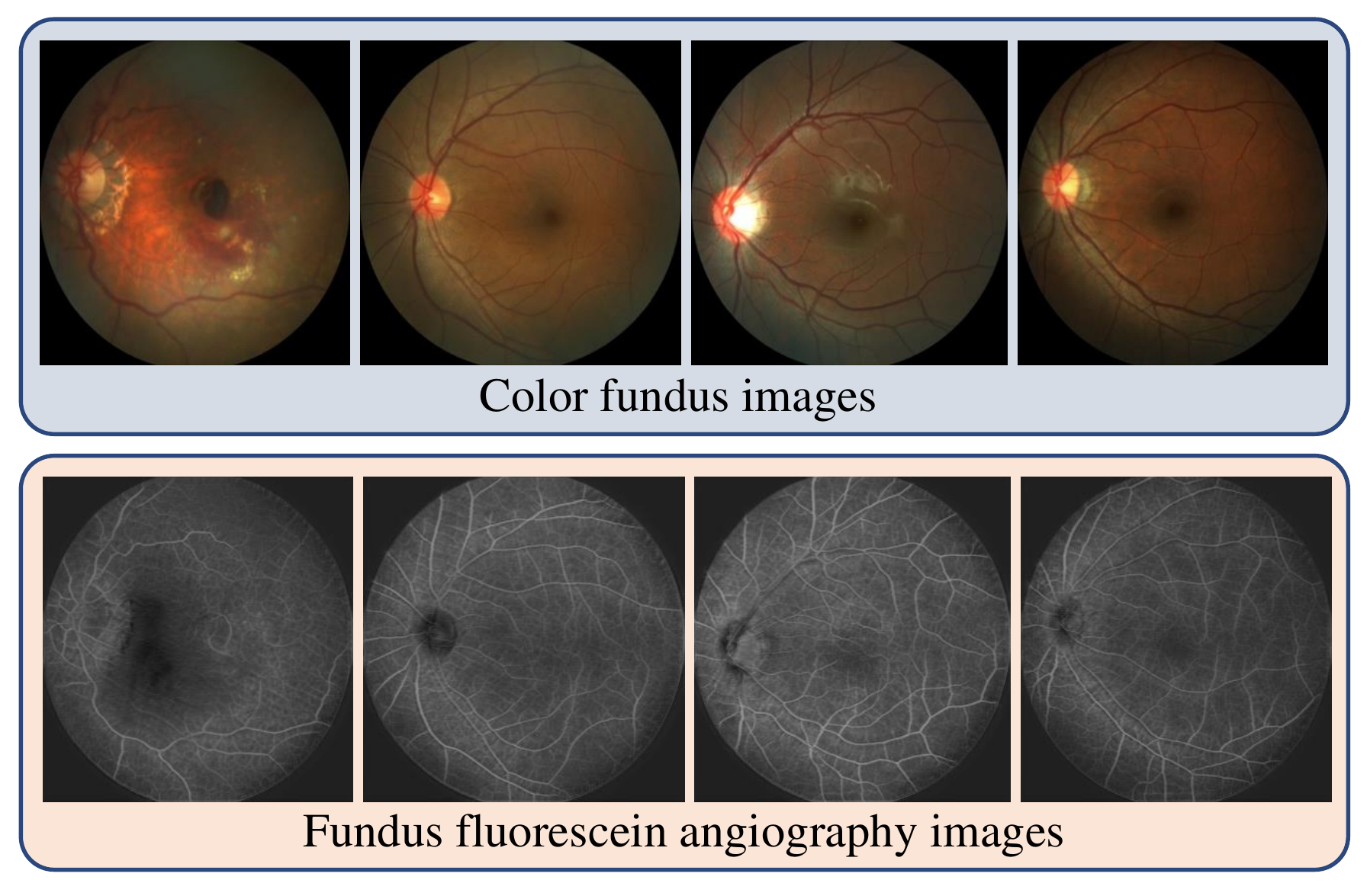}
	\caption{Examples of color fundus images and the corresponding synthesized FFA images. Fundus images are selected from the Ichallenge-AMD dataset [26], and FFA images are synthesized by training a CycleGAN~\cite{zhu2017unpaired} on a public Fundus-FFA dataset dataset~\cite{hajeb2012diabetic}, and then tested on the unseen fundus images. Our goal is to perform self-supervised learning using
		data from these two related modalities.}
	\label{fig:introdu}
\end{figure}

The self-supervised feature learning methods have been explored in the medical imaging domain for several tasks, such as subject identification from spinal MRI~\cite{jamaludin2017self}, cardiac MR image segmentation~\cite{bai2019self}, brain hemorrhage classification~\cite{zhuang2019self} and lung lobe segmentation and nodule detection task~\cite{tajbakhsh2019surrogate}.
Most previous works focused on developing novel pretext tasks as the supervisory signals to train the network to learn feature representation. 
For example,~\citet{bai2019self} proposed to learn self-supervised features by predicting anatomical positions from MR images.  
\citet{zhuang2019self} designed a new pretext task, \ie, Rubik's cube recovery, as a supervisory signal to train the network to predict these transformations. 
The common idea of these works is to exploit internal structures of data and encourages the network to predict such structures. 
However, most previous works are focused on learning self-supervised features with single modality data, while none of them investigate the role of multi-modal data and how it could be utilized in self-supervised learning.

Fundus fluorescein angiography (FFA) is an imaging modality that can provide useful information regarding the retinal vasculature in the retina~\cite{patel2015color}.
This information can help ophthalmologists better understand the structures of fundus lesions, microangioma, and capillary non-perfusion area, which are crucial for the diagnosis and treatment of some vitreoretinal diseases like AMD and PM~\cite{wang2017fundus,patel2015color,hoang2020imaging}.
The retinal vasculature information presented in FFA is complementary to color fundus images since FFA could identify fundus lesions that were not discovered by color fundus images~\cite{wang2017fundus}. 
Moreover, as shown in~\cite{patel2015color}, compared to using color fundus photographs alone, there is a significant improvement in diagnostic sensitivity when using color fundus photographs with the corresponding FFA images.
To utilize the mutual information in these two modalities, we propose to learn \emph{the general feature representations for fundus disease classification via both color fundus and the corresponding FFA images.}

However, FFA is an invasive and time-consuming procedure, which is difficult to collect in many clinical sites. 
To the best of our knowledge, the Fundus-FFA dataset~\cite{hajeb2012diabetic} is the only publicly available color fundus images with the corresponding FFA.
\revise{Hence, we obtain the FFA modality through a generative adversarial network on the public fundus-FFA dataset~\cite{hajeb2012diabetic}, such that our method is still applicable even when only color fundus images are available.}
Some examples of the color fundus and the corresponding synthesized FFA images are shown in Figure~\ref{fig:introdu}. 
Naively concatenating more datasets is a simple solution to utilize multiple modalities. However, as has been discovered in supervised learning, \revise{it is not an efficient way to use multiple modalities}, which can even lead to reduced performance on the dataset of interest. 
We also observed this phenomenon for self-supervised learning, as results in Figure~\ref{fig:feature}(a) in Section IV. 
By enlarging the dataset with corresponding FFA images, self-supervised learning performance decreases by around 6\%.   
This fact suggests that the presence of domain difference can affect the performance and cross-modality relationship has to be considered.



To address this issue, in this paper, we formulate a novel patient feature-based softmax embedding to learn general feature representation from multi-modal data by learning the \emph{positive concentrated} and \emph{negative separated} properties.
\emph{The positive concentrated property} refers to learn \textbf{transformation-invariant and modality-invariant features} for individual patients. 
This is motivated by the fact that our downstream task is disease classification and a patient's disease classification result would not change due to image transformations, thus the expected feature representation should be invariant to transformations. 
Similarly, a patient's two modalities,~\ie, a color fundus image and the corresponding FFA image, should share the same semantic meaning, thus their feature representations should be coherent. 
Hence, we propose to mine the shared cross-modality information by learning modality-invariant features.~\revise{
	\emph{The negative separated property} refers to learn \textbf{patient-similarity features} by separating patients from each other. 
	This is based on the observation that class-wise supervised learning can retain  apparent similarity among classes in the representation space.
	For an image from a class leopard, the classes that get the highest responses from a trained neural net classifier are all visually correlated, e.g., jaguar and cheetah. It is not the semantic labeling, but the apparent similarity in the data themselves that brings some classes closer than others. 
	Hence, if we treat each patient as a class and learn to separate him/her from others, we may end up with a representation that captures apparent similarity among patients.} 
With these constraints as our learning objective, the network encodes both~\emph{modality-invariant} features and \emph{patient-similarity} features into high-level representations, which can capture the semantically shared information across different modalities and apparent visual similarity between patients; see demonstrations in Figure~\ref{fig:feature} and Figure~\ref{fig:knn}. 
\revise{A ``patient'' is a triplet, consisting of color fundus, transformed fundus, and FFA images, where all these images are obtained from the same patient. ``patient feature-based'' denotes that our loss function is calculated directly based on the patients’ features. Hence, we name our method as “patient feature-based softmax embedding”.} 

To demonstrate the effectiveness of our method, we employed two public fundus image datasets for disease classification, \ie, Ichallenge-AMD dataset~\cite{2020adam}, and Ichallenge-PM dataset~\cite{2020paml}. 
Given that the proposed self-supervised method does not use any label information, a direct comparison between our method and
the state-of-the-art retinal disease diagnosis methods might not be
fair, but extensive experiments still demonstrate the superiority of our method against the state-of-the-art self-supervised methods~\cite{wu2018unsupervised, chen2020simple,ye2019unsupervised} on two retinal disease datasets. 
Notably, our method also achieves comparable performance with the supervised baseline. 

The main contributions of this paper are:

\begin{itemize}
	\item We present a novel self-supervised learning method by effectively exploiting multi-modal data for retinal disease diagnosis.
	Our method contains a network to synthesize another modality, thus it is still applicable even though only color fundus images are available.  
	To the best of our knowledge, this is the first work for self-supervised disease diagnosis from fundus images. 
	
	\item We formulate the patient feature-based softmax embedding as a self-supervised signal to capture the mutual information across modalities and patient-similarity features from multi-modal data, which learns effective representation for fundus image classification. 
	
	\item 
	Extensive experiments on two common eye diseases, \ie, AMD and PM, demonstrate the superiority  of our method than other state-of-the-art self-supervised methods.
	Our method also achieves comparable results with the supervised baseline.  
	
\end{itemize}

\section{Related Works}
\begin{figure*}[!t]
	\centering
	\includegraphics[width=0.85\textwidth]{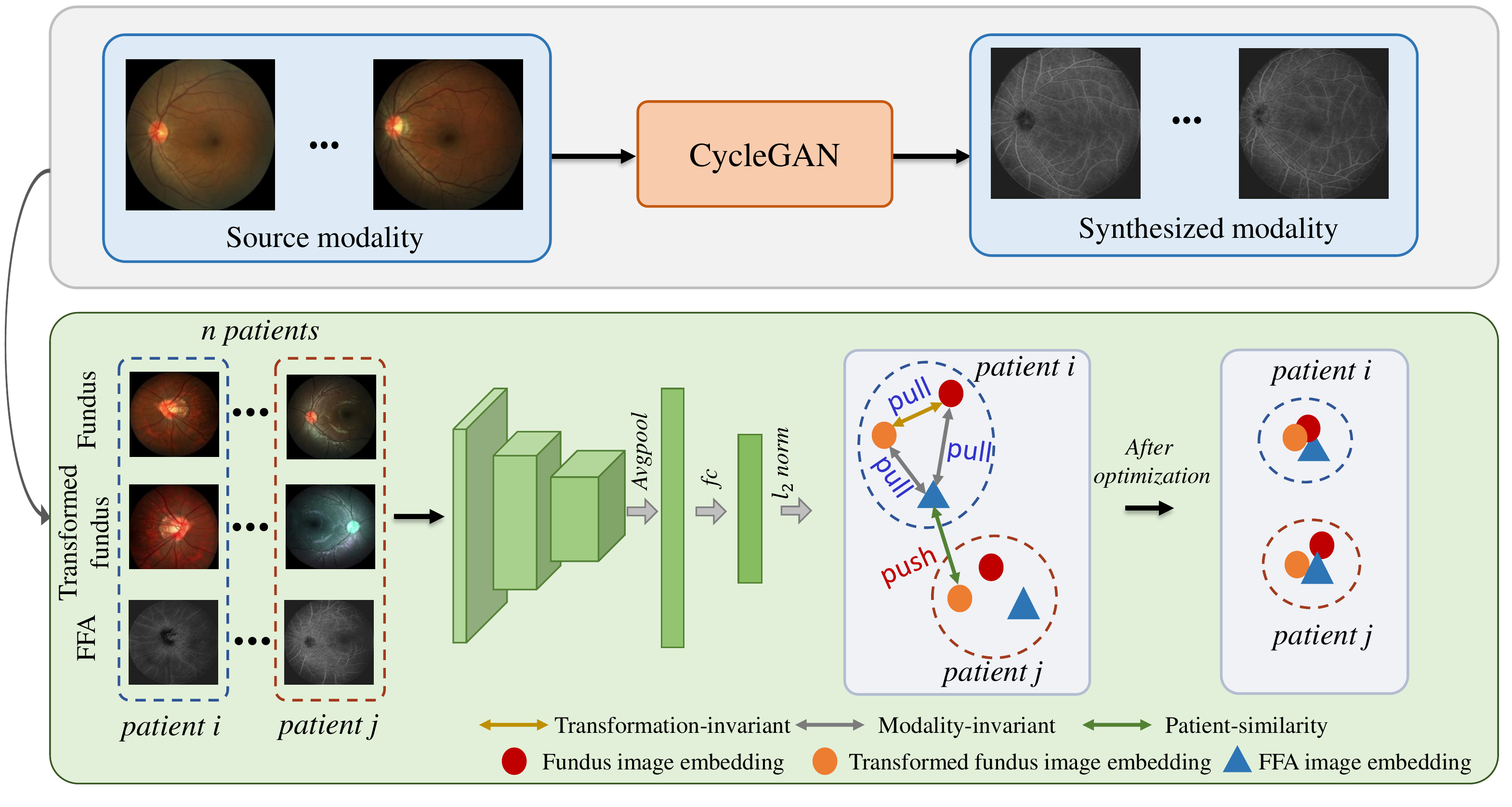}
	\caption{The illustration of the proposed method. We first train a generative network (CycleGAN) on the fundus-FFA dataset to learn the mapping function between color fundus image and FFA, and then synthesize the corresponding FFA on target color fundus datasets. Secondly, triplets are derived from each patient, consisting of the randomly selected fundus image, transformed image, and the corresponding FFA.    
		These triplets are fed into the neural network to learn the high-level representation with our proposed patient feature-based softmax embedding objective. Finally, \revise{the network is evaluated on} unseen fundus images and the final classification result is obtained by applying a KNN on the features.}
	\label{fig:pipeline}
\end{figure*}
In this section, we mainly review automatic disease diagnosis works from fundus photography and literatures related to self-supervised feature learning.

\subsection{Automatic Disease Diagnosis from Fundus Photography}
Many conventional ophthalmic diseases can be examined from fundus photography, such as age-related macular degeneration (AMD), diabetic retinopathy (DR), glaucoma,
and pathological myopia (PM). 
With the advances of deep learning, considerable efforts have been devoted to developing convolutional neural networks for automatic eye disease recognitions~\cite{fu2018disc,burlina2016detection,li2019canet,yin2019pm,cheng2019structure,milea2020artificial,grassmann2018deep,peng2019deepseenet,freire2020automatic}. As for AMD diagnosis,~\citet{burlina2016detection} employed a CNN that is pre-trained on OverFeat features to perform AMD classification from fundus images.~\citet{grassmann2018deep} ensembled several convolutional neural networks to classify AMD diseases into 13 classes.~\citet{peng2019deepseenet} developed a DeepSeeNet based on an Inception-v3 architecture~\cite{xia2017inception} to identify patient-level AMD severity, by first detecting individual risk factors and then combining values from both eyes to assign a severity result. As for PM classification,~\citet{freire2020automatic} employed Xception~\cite{chollet2017xception} as the baseline architecture with ImageNet pre-train weights to diagnose PM from fundus images. 
However, most previous works on disease diagnosis from fundus photography are based on supervised learning, which requires a massive amount of labeled data. 
Different from previous works, in this paper, we focus on developing the self-supervised method for fundus disease diagnosis.

\subsection{Self-supervised Feature Learning}
Self-supervised feature learning is becoming a popular topic and has been studied in several medical image recognition tasks.  
The common principle of existing works is to construct different pretext tasks by discovering supervisory signals directly from the input data itself, and then training the deep network to predict this supervisory information, such that the high-level representation of the input is learned. 
Notable pretext tasks include Rubik's cube recovery~\cite{zhuang2019self}, anatomical position prediction~\cite{bai2019self}, reconstructing part of the image like image completion~\cite{zhou2019models, chen2019self}, 3D distance prediction~\cite{spitzer2018improving}, image-intrinsic spatial offset prediction~\cite{blendowski2019learn}. 
For example,~\citet{zhuang2019self} proposed Rubik’s cube
recovery task for brain hemorrhage classification and tumor segmentation from CT and MR images, respectively.~\citet{bai2019self} proposed to learn self-supervised features by predicting anatomical positions for cardiac MR image segmentation. 
\citet{blendowski2019learn} designed image-intrinsic spatial offset relations task to learn self-supervised features. 
\citet{spitzer2018improving} introduced predicting 3D distance between two patches sampled from the same brain as a pretext task.

Recently, instance discrimination~\cite{he2019momentum,wu2018unsupervised,ye2019unsupervised}, an effective pretext task, achieves promising results on unsupervised feature representation learning. 
For example,~\citet{dosovitskiy2015discriminative} proposed to use softmax embedding with classifier weights to calculate the feature similarity, however, it prevents explicitly comparison over features, which results in limited efficiency and discriminability.
\citet{wu2018unsupervised} developed memory bank to memorizes features of each instance. 
\citet{ye2019unsupervised} calculated the positive concentrated property based on the ``real'' instance feature, instead of classifier weights~\cite{dosovitskiy2015discriminative} or memory bank~\cite{wu2018unsupervised}.  
However, this method treated the optimization as the binary classification problem via maximum likelihood estimation, which is infeasible to learn the feature embedding from multi-modal data. 

Unlike the previous works that explored self-supervised learning from the single modality data, we present to effectively exploit multi-modal data to improve self-supervised feature learning.
\revise{Multi-modality data has been widely utilized for many medical image recognition tasks and there are several related works \cite{li2016multi,brand2019joint,zhou2020hi,zhou2019latent,li20183d,xing2020artificial}. 
For example,~\citet{brand2019joint} proposed to identify Alzheimer’s disease (AD) - relevant
biomarkers by learning from two modalities, \ie, genetic information and brain scans.
\citet{zhou2020hi} developed a hybrid fusion network for multi-modal MR image synthesis. 
~\citet{zhou2019latent} proposed a latent representation learning method for
multi-modality (\ie, PET and MRI) based AD diagnosis. 
However, all these works are supervised learning or image synthesis methods, while we are investigating an unsupervised learning strategy for disease classification. } 

From this perspective, in this work, we employ the instance discrimination~\cite{wu2018unsupervised,ye2019unsupervised} as the pretext task, and propose to learn features by learning both modality-invariant and patient-similarity features from multi-modal data.

\section{Method}

\subsection{Overview}
Figure~\ref{fig:pipeline} depicts the workflow of our self-supervised method for retinal disease diagnosis.
We first train a GAN model on the Fundus-FFA dataset~\cite{hajeb2012diabetic} to learn the mapping function between the color fundus and FFA, and then obtain the synthesized FFA modality on the Ichallenge-AMD and Ichallenge-PM dataset.
Secondly, to learn the self-supervised features, we randomly sample $n$ triplets, and each triplet is derived from each patient, consisting of color fundus image, the transformed image, and the corresponding FFA. 
The triplets are fed into the neural network to learn the high-level feature representations, which are optimized by the proposed patient feature-based softmax embedding objective. 
Our learning objective encourages the network to learn transformation- and modality-invariant features, while also capture the patient-similarity features. 
Finally, we evaluate the network on unseen fundus images, following the standard evaluation protocol in most self-supervised works~\cite{he2019momentum,wu2018unsupervised}. 
The final classification result is obtained by applying a K-Nearest Neighbor (KNN) classifier. 
Below, we will elaborate on the FFA image synthesization, patient feature-based softmax embedding, and technical details.



\subsection{FFA Image Synthesization} 

FFA is invasive and it is difficult to collect in many clinical sites~\cite{jayadev2016utility}. 
Hence, we propose to synthesize FFA images, such that our method can still be utilized to perform self-supervised learning even though only color fundus are available.
Specifically, we train a generative model on the Fundus-FFA dataset~\cite{hajeb2012diabetic} to learn the mapping function between the color fundus and FFA images, and then synthesize the corresponding FFA modality in the target fundus datasets, \ie, Ichallenge-AMD and Ichallenge-PM dataset, to perform self-supervised feature learning.

The Fundus-FFA dataset~\cite{hajeb2012diabetic} contains color fundus images with the corresponding FFA images, which is not pixel-aligned. Based on this consideration, we trained a CycleGAN model~\cite{zhu2017unpaired} and we followed the original setting to train the network with both adversarial loss and cycle-consistency loss. 
To adapt the network to our task, we modified the learning rate to 0.0001 and trained it for 500 epochs. After network optimization, we tested the model on unseen fundus datasets, and the synthesized results can be seen in Figure~\ref{fig:introdu} and Figure~\ref{fig:result_classify}. 
Since there is no ground-truth FFA provided in these datasets, the synthesization quality is measured by running supervised learning for fundus image classification; see results in Table~\ref{tab:ffa}.

\subsection{Patient Feature-based Softmax Embedding}

Let  $ C  = \left \{ c_i  \right \} $ and $ S = \left \{ s_i  \right  \}$, where $c_i $ and $ s_i$ denote the color fundus image and the corresponding FFA image of patient $i$, respectively. 
Our goal is to learn a feature embedding network $f_\theta ( \cdot )$ that maps an unlabeled image $c_i$ or $s_i$ to a low-dimensional feature embedding $f_\theta( c_i) $ or  $ f_\theta (s_i) \in \mathbb{R}^{d}$, where $d$ is the feature dimension.
For simplicity, we use ${\bf f}_i = f_\theta(c_i)$, ${\bf g}_i = f _\theta(s_i) $ to represent the feature of patient $i$ from fundus and FFA, respectively.
We normalize all the features by $l_2$ normalization, \ie, $ \left \| {\bf f}_i  \right \|_2 = 1$, $ \left \| {\bf g}_i  \right \|_2 = 1$.
Without the ground-truth category labels, we need to form self-supervised learning constraints to facilitate model optimization.

\begin{figure}[t]
	\centering
	\includegraphics[width=0.5\textwidth]{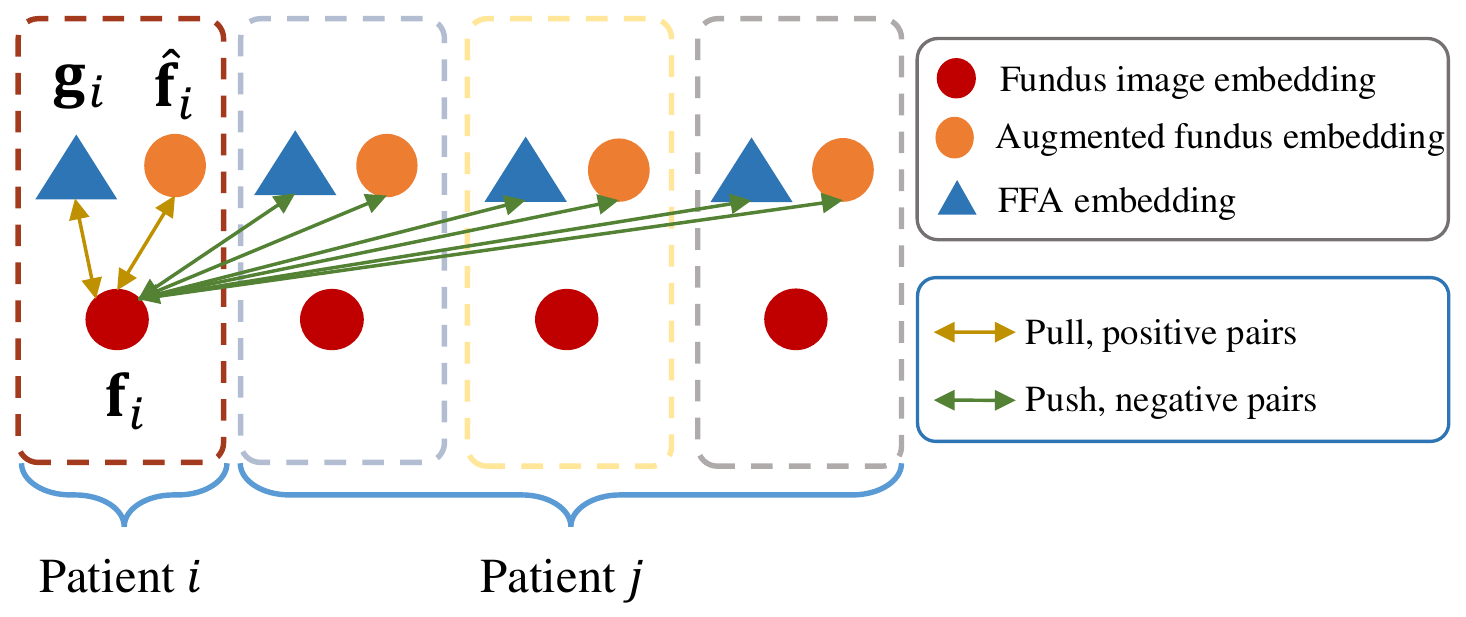}
	\caption{\revise{The illustration of the proposed patient-based softmax feature objective.}} 
	\label{fig:method}
\end{figure}

\para{Transformation- and modality-invariant features.}
A patient's disease diagnosis result would not change due to image transformations. Thus, a good feature embedding should satisfy that the representation of a color fundus image $c_i$ of patient $i$ should be invariant under random data augmentations.  
Intuitively, a color fundus image $c_i$ from a patient $i$ should share the same semantic meaning with the corresponding FFA image $s_i$, thus the representations of a patient should be coherent. Hence, the network requires to learn \emph{transformation-} and \emph{modality-invariant} features. 
%
To achieve this, we randomly sample $n$ patients from the datasets, and each patient consists of both color fundus image and the corresponding synthesized FFA. The selected samples are denoted by $\left \{c_1, s_1, \cdots, c_n, s_n \right \}$. 
To learn the transformation-invariant features, a random data augmentation is applied to slightly modify the original fundus $c_i$ to $\hat{c}_i$, and we can obtain the batch denoted by  $\left \{c_1, \hat{c}_1, s_1, \cdots, c_n, \hat{c}_n, s_n \right \}$. These images are fed into the network to get high-level representations, \ie, $\left \{{\bf f}_1, \hat{{\bf f}}_1, {\bf g}_1, \cdots, {\bf f}_n, \hat{{\bf f}}_n, {\bf g}_n \right \}$. 
The probability of $\hat{c}_i$ being recognized as patient $i$ is defined as 
\begin{equation}
P(i| {\hat{c}}_{i}) = \frac{   {\rm exp} \left ({ {\bf f}^{T}_i} {\bf  \hat{f}}_{i}/\tau \right )   }{\sum_{k=1}^{n} {{\rm exp} \left ( {\bf f}_{k}^{T} {\bf  \hat{f}}_{i}/\tau  \right ) }},
\label{eq:1}
\end{equation}
where $\tau$ is the temperature parameter controlling the concentration level of the sample distribution~\cite{hinton2015distilling}. 
The probability of $s_i$ being recognized as patient $i$ is defined by 
\begin{equation}
P(i| {s}_{i}) = \frac{   {\rm exp} \left ({ {\bf f}^{T}_{i}} {\bf  g}_{i}   /\tau \right )   }{\sum_{k=1}^{n} {{\rm exp} \left ( {\bf f}_{k}^{T} {\bf  g}_{i}/\tau  \right ) }},
\label{eq:syn}
\end{equation}
where $ {\bf f}^{T}_i {\bf  \hat{f}}_{i} $, $ {\bf f}^{T}_i {\bf g}_{i} $ denote the cosine similarity between positive pairs, as shown in Figure~\ref{fig:method}. 

\para{Patient-similarity features.}
\if 
A patient is distinctive in its own right, and the provided fundus and corresponding FFA image could differ significantly from images of other patients.
This follow the same intuition as the instance discrimination task~\cite{wu2018unsupervised}. 
Hence, a good feature embedding should also satisfy the different patients' images should be dissimilar in the representation space. 
\fi 
To learn patient-similarity features, we need to treat each patient as a class and learn to separate him/her from other patients. 
As shown in Figure~\ref{fig:method}, the distance between negative pairs should be enlarged in the representation space. 
The probability of ${c}_j$ being recognized as patient $i$ is defined by
\begin{equation}
P(i| c_j) = \frac{   {\rm exp} \left ({ {\bf f}^{T}_{i}} {\bf  f}_{j}/\tau \right )   }{\sum_{k=1}^{n} {{\rm exp} \left ( {\bf f}_{k}^{T} {\bf  f}_{j}/\tau  \right ) }}, j \neq i.
\label{eq:3}
\end{equation}
This equation also holds for $s_j$. 
We assume different image samples being recognized as patient $i$ are independent, the joint probability of $ \hat{c}_i, s_i $ being recognized as patient $i$ and $c_j, s_j$  being not classified to patient $i$ is  
\begin{equation}
P_i = P(i| \hat{c}_i) P(i| {s}_{i}) \prod_{j \neq i} (1 - P(i | c_j)) \prod_{j \neq i} (1 - P(i | s_j)),
\end{equation}

\para{Learning objective.}
The above probability is optimized by the negative log likelihood, which is defined as 
\begin{equation}
\begin{split}
\mathcal{L}_i = &- {\rm log }{P(i | \hat{c}_i)} -  {\rm log }{P(i | s_i)}  - \sum_{j \neq i}{ {\rm log}(1 - P (i|c_j))}  \\ & - \sum_{j \neq i}{ {\rm log}(1 - P (i|s_j))}
\end{split}
\label{eq:loss_d}
\end{equation}
The final loss function is defined by minimizing the mean of the negative log likelihood over all patients $n$ within the batch.
Our learning objective is defined as
\begin{equation}
\begin{split}
\mathcal{L} = \frac{1}{n} \sum _{i} \mathcal{L}_i. 
\end{split}
\label{eq:loss_total}
\end{equation}
Hence, we learn the \emph{modality-invariant} and \emph{patient-similarity} features by simultaneously doing positive concentration and negative separation.  
Since the loss function is calculated on the feature of the patient, we name it as \emph{patient feature-based softmax embedding.}

\subsection{Technical Details}
\para{Network architecture.}
Our framework is based on the ResNet18 backbone~\cite{he2016deep}, following the same setting as the previous works~\cite{wu2018unsupervised,ye2019unsupervised}.
We apply an average pooling on the output of the last residual block in ResNet18. Then, the feature is flattened to a vector and a fully connected layer, a batch normalization layer, and a ReLU are sequentially applied to reduce the feature dimension to 128.  Finally, the feature is normalized by $l_2$ normalization to the embedding space. 
The proposed patient feature-based softmax embedding loss function is utilized to train the neural network. 

\if 1 
\para{Training strategies.}
At each training iteration, $m$ images are randomly selected, and random data augmentation are applied twice to the selected instances, resulting in $2m$ generated image instances. 
Then each image is rotated by $\left \{  0^\circ, 90^\circ, 180^\circ, 270^\circ \right \} $ to derive the
rotation-transformed instances. The final batch size is $8m$.
This training strategy also takes the full advantage of relationships among all instances sampled in a mini-batch. 
To evaluate the learned feature, we apply k-nearest neighbors classifier based on the $\bf f$ and $k$ is set to 100. 
\fi

\para{Implementation details.}
The whole framework is built on PyTorch~\cite{paszke2017automatic} with an NVIDIA Tesla V100 32GB GPU. 
We resize images to $320 \times 320$ resolution. For data augmentation, we randomly scale and crop images into the patches of size $224 \times 224$, with a random scaling factor chosen from [0.2, 1.0].  \revise{Our algorithm performs randomly horizontal flip and has a probability of 0.2 to randomly grayscale the input. The algorithm also randomly blends the image to some extent with its black version, grayscale version. This operation changes the brightness, contrast, and saturation of the input image with a random factor chosen uniformly from [0.6, 1.4], following the setting in~\cite{ye2019unsupervised,wu2018unsupervised}.} 
Note that data augmentation is also applied in each image in the triplet to enrich the training samples. 
For implementation, each image has two positive samples and $2n-2$ negative samples to compute Eq.~(\ref{eq:loss_total}), where $n$ is the number of sampled patients and $\tau$ is set to 0.1.
In each feed forward, we sample $75$ patients, \ie, $n=75$.    
The network is optimized with Adam optimizer~\cite{kingma2014adam}. The initial learning rate is set to 0.0001 and is dropped by a factor of 0.1 every 1000 epochs.  
All the experiments are equally trained for 2000 epochs and the reported results are conducted on 5-fold cross-validation.

\revise{\para{Evaluation protocol.}}
We verify our method by applying a KNN classifier on frozen features, following a common protocol~\cite{wu2018unsupervised,ye2019unsupervised}.
To investigate the transfer learning capacity, we unfreeze the features and train a supervised linear
classifier (a fully-connected layer followed by softmax) on the target datasets.

\section{Experiments}

\subsection{Datasets}
We employ two public retinal disease datasets,~\emph{i.e.}, 
Ichallenge-AMD~\footnote{https://ichallenges.grand-challenge.org/iChallenge-AMD/} (task 1) and Ichallenge-PM~\footnote{https://ichallenges.grand-challenge.org/iChallenge-PM/} (task 1), and evaluate the effectiveness of our method by performing normal and abnormal fundus image classification. 

\para{Ichallenge-AMD dataset.}
Ichallenge-AMD dataset~\cite{2020adam} contains 1200 annotated retinal fundus images, including both non-AMD subjects (77\%) and AMD patients (23\%). Typical signs of AMD that can be found in these photos include drusen, exudation, hemorrhage, etc. 
Since only training data is released with annotations, we use the training data in the Ichallenge-AMD dataset and perform 5-fold cross-validation.

\para{Ichallenge-PM dataset.}
Ichallenge-PM dataset~\cite{2020paml} contains 1200 annotated color fundus photos with Non-PM (50\%) and PM cases (50\%). 
All the photos were captured with Zeiss Visucam 500. 
We use the training data in the Ichallenge-PM dataset and perform 5-fold cross-validation. 
In these two datasets, the image-level annotation is provided, where 0 denotes normal and 1 denotes abnormal cases. 
However, we do not utilize any human-annotated labels information during network training. 
To evaluate the classification accuracy of our method, we employ AUC, accuracy, precision, recall, and F1-score as the evaluation metrics.

\para{EyePACS dataset.}
To evaluate the transfer learning capacity of our model, we train the self-supervised model on the Kaggle’s Diabetic Retinopathy Detection Challenge (EyePACS) dataset~\footnote{https://www.kaggle.com/c/diabetic-retinopathy-detection/data} and report the classification result on the AMD dataset. 
This dataset is sponsored by the California Healthcare Foundation. It provides a totally 88,702 images, captured under various conditions and various devices. The Left and right fields are provided for every subject, and an ophthalmologist rated the presence of diabetic retinopathy in each image on a scale of 0 to 4. We use all the images in this dataset to train our self-supervised model. 
Note that we did not utilize any human-annotated labels in this dataset.

\para{Fundus-FFA dataset.}
To the best of our knowledge, \revise{Fundus-FFA dataset~\cite{hajeb2012diabetic} is the only publicly available dataset that contains color fundus images and corresponding FFA images. It has 30 healthy persons and 29 patients with diabetic retinopathy. Each patient has a color fundus photo and corresponding FFA}. The dataset is very limited and is not suitable to train an unsupervised model. As an alternative, we train a generative model on the Fundus-FFA dataset to learn the mapping function between the color fundus images and the corresponding FFA images~\cite{costa2017end, zhu2017unpaired,hervella2019paired,li2019unsupervised}. 

\subsection{Comparison  on the Ichallenge-AMD Dataset}
To show the effectiveness of our method, we compare it with state-of-the-art self-supervised learning methods on the Ichallenge-AMD dataset.

\revise{\para{Experimental settings.}} 
To have a fair comparison, all of the models are trained on the ResNet18 backbone~\cite{he2016deep} with 5-fold cross-validation. In the \emph{Supervised} baseline, we modified the output channel of the original fully connected layer of ResNet18 to 2 for two-class classification. The supervised model is trained by the cross-entropy loss with human-annotated labels. 
To compare with self-supervised methods, unlike other self-supervised methods that learn the 2D or 3D correspondences by predicting rotation or anatomical positions~\cite{bai2019self,zhuang2019self}, our fundus image classification task is invariant to the image transformation.
To the best of our knowledge, there are no related self-supervised methods that learn self-supervised transformation-invariant features in the medical imaging domain, we compare with several state-of-the-art instance discrimination methods in the computer vision domain~\cite{wu2018unsupervised,chen2020simple,ye2019unsupervised}.
Finally, we perform KNN on all the unsupervised feature learning methods to evaluate the feature performance for classification and $k$ = 100.
Note that we run these methods with the same backbone, learning strategies and trained all the models for 2000 epochs on 5-fold cross-validation.

\begin{figure}[!t]
	\captionsetup[subfloat]{font=Large,labelfont=Large}
	\resizebox{0.5\textwidth}{!}{
		\centering
		\subfloat[Ichallenge-AMD dataset]{\includegraphics[scale=0.5]{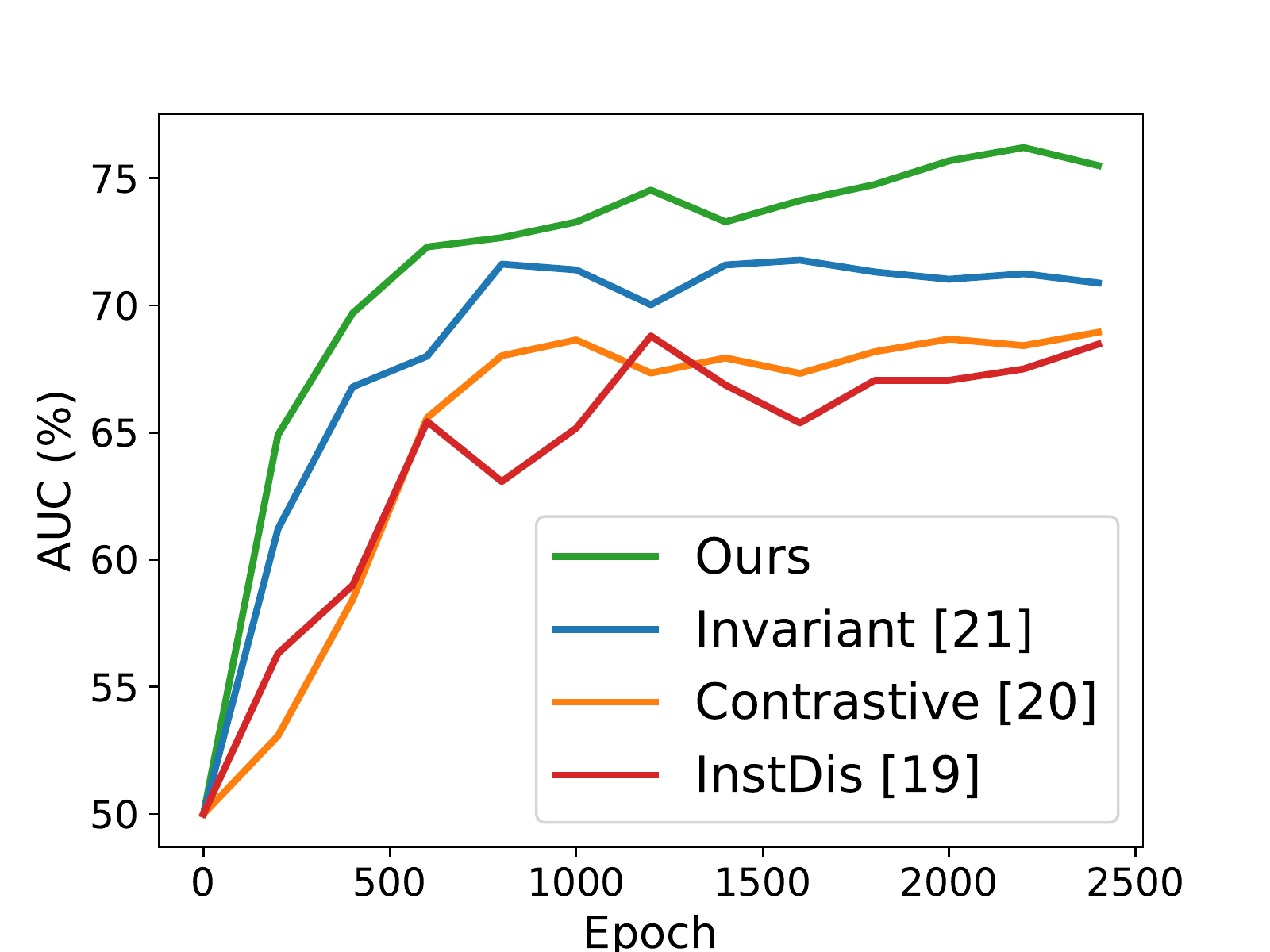}}
		\subfloat[Ichallenge-PM dataset]{\includegraphics[scale=0.5]{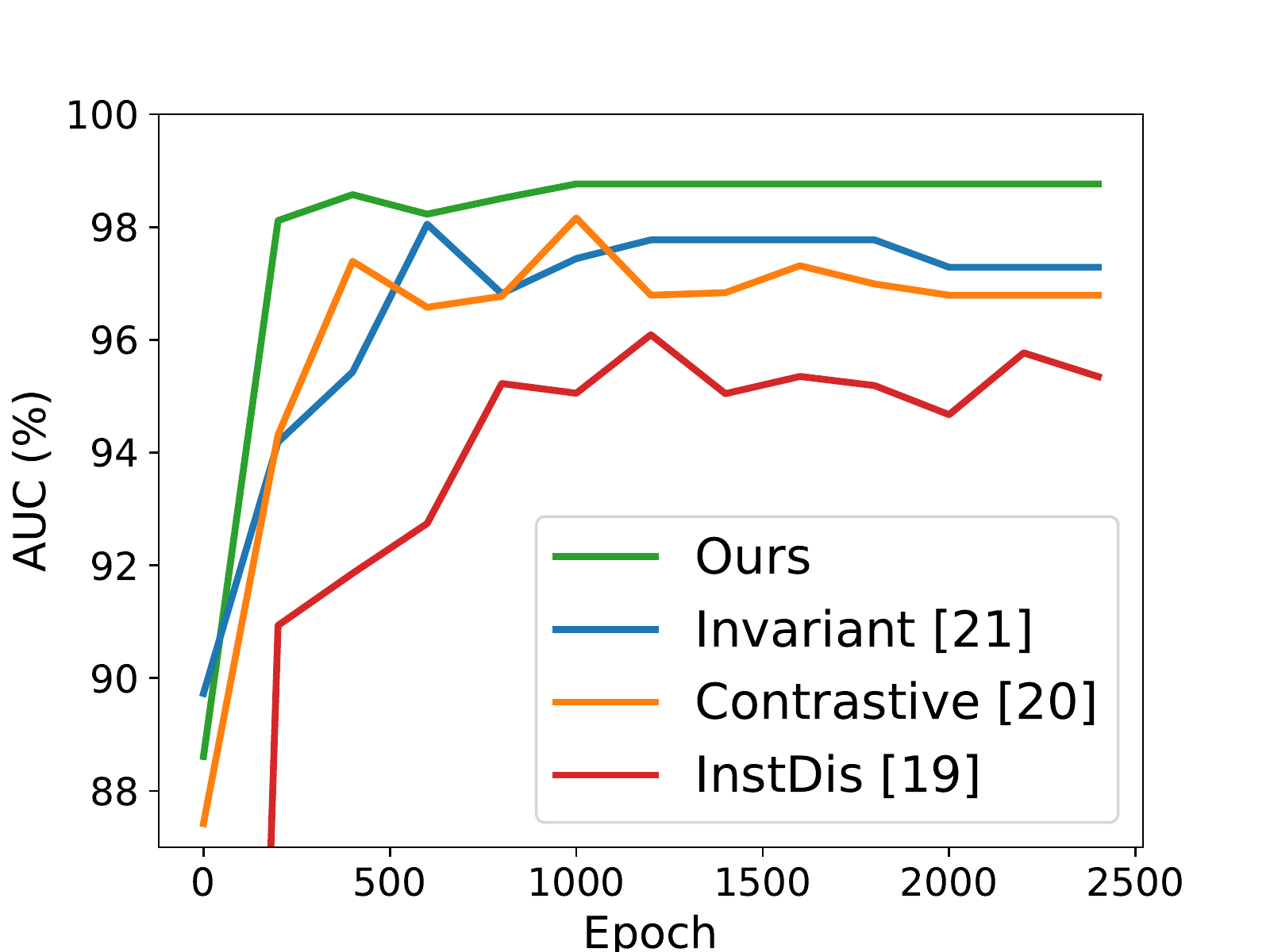}}}
	\protect\caption{Comparison of AUC results on the  (a) Ichallenge-AMD dataset and (b) Ichallenge-PM dataset. }\label{fig:auc_curve} 
\end{figure}

\para{Results.}
The results of different unsupervised methods are shown in Table~\ref{tab:result_amd}.
It is observed that \emph{Contrastive}~\cite{chen2020simple} and \emph{Invariant}~\cite{ye2019unsupervised} achieve better results than \emph{InstDisc}~\cite{wu2018unsupervised}, showing that contrastive learning can be beneficial to unsupervised feature learning. 
From the comparison, we can see that \emph{Invariant}~\cite{ye2019unsupervised} performs slightly better than \emph{Contrastive}~\cite{chen2020simple}. This is because the heavy data augmentation proposed in \emph{Contrastive}~\cite{chen2020simple} would hurt the performance in our fundus image classification task. 
It is also observed in Table~\ref{tab:result_amd} that our method excels all other unsupervised feature learning methods by at least around 3.16\% on AUC, which demonstrates the effectiveness of our method in the unsupervised feature learning.  
Figure~\ref{fig:auc_curve}(a) visualizes the learning curve of the validation results and we can see our method consistently outperforms other methods. 
Notably, without any annotation during training, our method is approaching to the supervised learning baseline, \eg, 74.58\% vs 77.19\% on AUC.
The results further demonstrate the effectiveness of our self-supervised learned features.

\begin{table}[t]
	\centering
	\caption{Results on the Ichallenge-AMD dataset (Unit: \%).}
	\resizebox{0.5\textwidth}{!}
	{		
		\begin{tabular}{c|ccccc}
			\toprule[1.5pt]
			& AUC & Accuracy & Precision & Recall & F1-score 
			\tabularnewline \hline 	Supervised	&	 77.19 & 87.09 & 82.98 & 77.82  & 79.27
			\tabularnewline \hline  
			InstDisc~\cite{wu2018unsupervised}  & 66.49 & 82.02  & 74.63  & 66.49  & 68.69
			\tabularnewline 
			Contrastive~\cite{chen2020simple}  & 68.06 & 82.45  & 73.48   & 68.06 & 69.84
			\tabularnewline  
			Invariant~\cite{ye2019unsupervised}  & 71.42  & 84.31  & 77.99  & 71.42  & 73.67
			\tabularnewline 
			\textbf{Ours}  & \textbf{74.58} &  \textbf{86.58} & \textbf{83.20} & \textbf{74.58} &	\textbf{77.33}			
			\tabularnewline
			\bottomrule[1.5pt]
		\end{tabular}

		\label{tab:result_amd}}
\end{table}

\begin{table}[!t]
	\centering
	\caption{Results on the Ichallenge-PM dataset (Unit: \%). }
	\resizebox{0.5\textwidth}{!}
	{\begin{tabular}{c|ccccc}
			\toprule[1.5pt]
			& AUC & Accuracy & Precision & Recall & F1-score 
			\tabularnewline \hline 				
			Supervised & 98.04 & 97.66  & 97.30 & 98.04  &  97.53 
			\tabularnewline \hline 
			
			InstDis~\cite{wu2018unsupervised} & 95.49 & 95.32 & 95.04   & 95.49  & 95.18
			\tabularnewline

			Contrastive~\cite{chen2020simple}  & 96.98 & 96.94 & 96.67  & 96.98 & 96.68 
			
			\tabularnewline
			Invariant~\cite{ye2019unsupervised}  & 97.26 & 97.30 & 97.11 & 97.26 & 97.16	
			\tabularnewline
			\textbf{Ours}  & \textbf{98.55} & \textbf{98.65} & \textbf{98.60} & \textbf{98.55} & \textbf{98.57} \tabularnewline 
			\bottomrule[1.5pt]
		\end{tabular}
		\label{tab:result_gon}
	}
\end{table}

\subsection{Comparison on the  Ichallenge-PM Dataset}
We also compare our method with the other unsupervised feature learning methods on the Ichallenge-PM dataset. In this dataset, we use the same experimental settings as those in the Ichallenge-AMD dataset. 
Table~\ref{tab:result_gon} summarizes the results of different methods on the Ichallenge-PM.
From Table~\ref{tab:result_gon} we can see that our method excels other methods on all five metrics. 
In particular, our method outperforms the state-of-the-art method \emph{Invariant}~\cite{ye2019unsupervised} by 1.29\% on AUC. 
It is observed that the results keep consistent with those on the Ichallenge-AMD dataset, showing the effectiveness and generalization of our method.
The validation results during learning are visualized in Figure~\ref{fig:auc_curve}(b). We can see that our method consistently surpasses other methods. 
It is worth mentioning that our method achieves higher results than the supervised upper bound in this dataset, which further demonstrates the effectiveness of our method.

\begin{table*}[t]
	\centering
	\caption{Results obtained by first training a self-supervised model on the EyePACS dataset and then fine-tuning on the following two datasets. \emph{Random init} denotes the network is trained with randomly weight initialization (Unit: \%). }
	
	{\begin{tabular}{c|ccccc|ccccc}
			\toprule[1.5pt]
			\multirow{2}{*}{\diagbox{Method}{Dataset}} & \multicolumn{5}{c|}{Ichallenge-AMD}& \multicolumn{5}{c}{Ichallenge-PM}
			\tabularnewline \cline{2-11} 
			& AUC & Accuracy & Precision & Recall & F1-score & AUC & Accuracy &  Precision & Recall & F1-score 
			\tabularnewline \hline 				
			Random init &  77.19 & 87.09 & 82.98 & 77.82  & 79.27  &  98.04 & 97.66  & 97.30 & 98.04  &  97.53  
			\tabularnewline
			
			Invariant~\cite{ye2019unsupervised}    & 81.62 & 87.51 & 81.92 & 81.62 & 81.35 & 98.02 & 97.84 & 97.56 & 98.02 & 97.75
			
			\tabularnewline 
			\textbf{Ours}    & \textbf{83.17} & \textbf{89.37} & \textbf{85.71} & \textbf{83.17} & \textbf{83.67} & \textbf{98.41} &\textbf{98.38} & \textbf{98.31} & \textbf{98.41} & \textbf{98.33} \tabularnewline 
			\bottomrule[1.5pt]
		\end{tabular}
		\label{tab:transfer}
	}
	
\end{table*}

\begin{figure*}[!t]
	\captionsetup[subfloat]{font=scriptsize,labelfont=scriptsize }
	\resizebox{1.0\textwidth}{!}{
		\centering
		\subfloat[Enlarged-Data: AUC 65.72\%]{\includegraphics[scale=0.3]{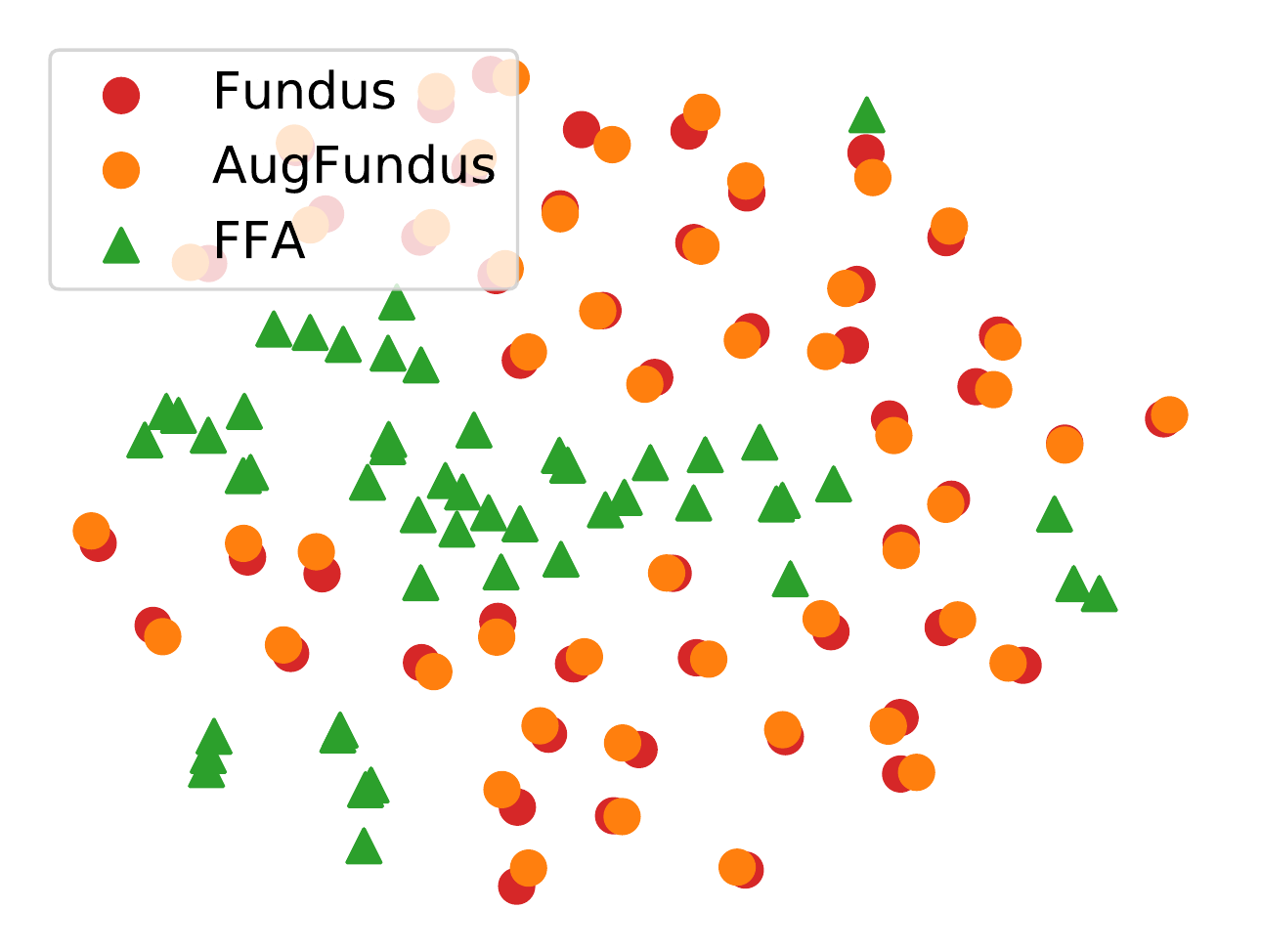}}
		\subfloat[As-Augmentation: AUC 70.93\%]{\includegraphics[scale=0.3]{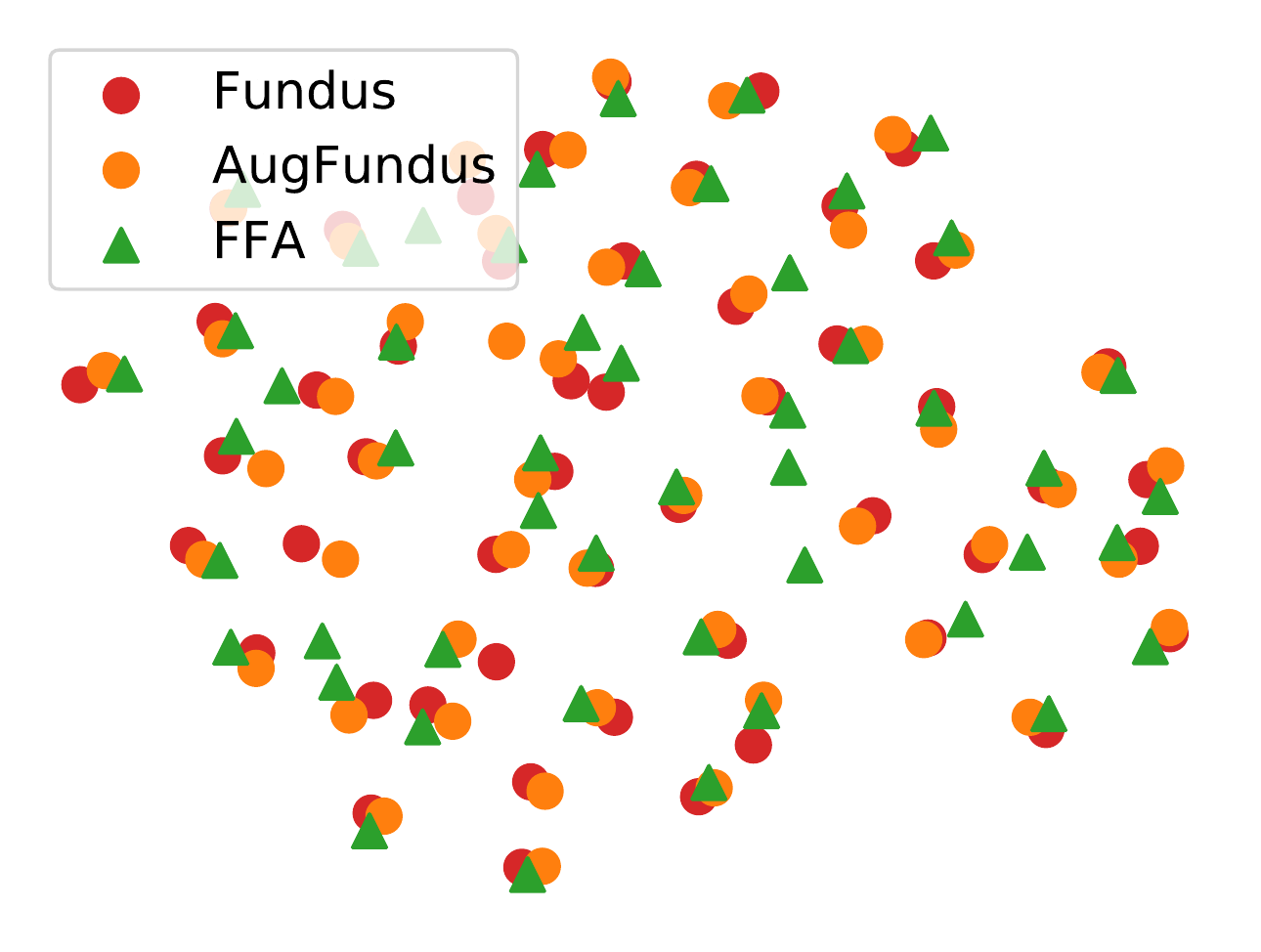}
		}
		\subfloat[Ours: AUC 74.58\%]{\includegraphics[scale=0.3]{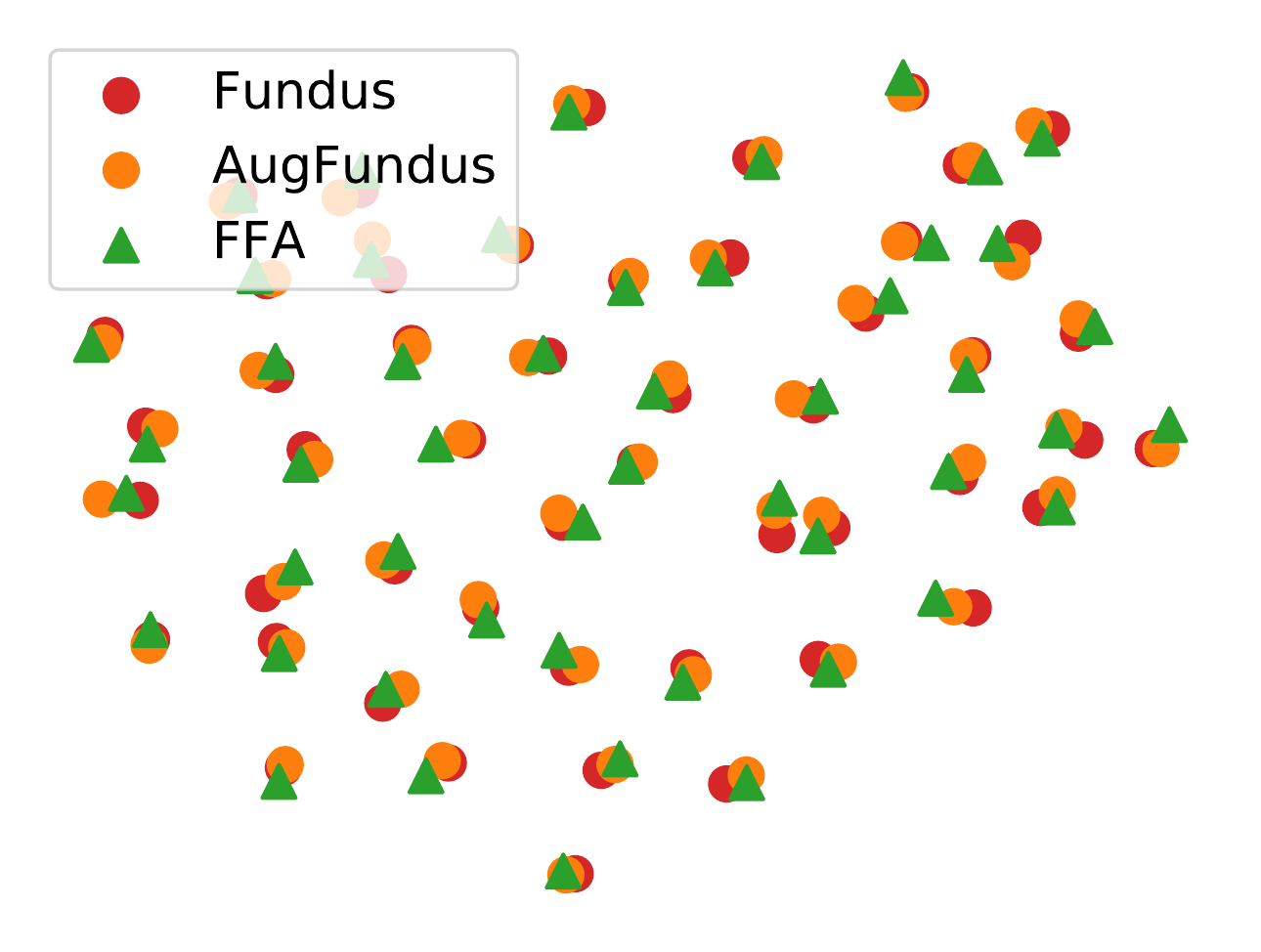}
		} }
		\protect\caption{Visualization of the learned feature embedding of three variants (see definitions in Table~\ref{tab:ablation_study}). 
			\revise{We randomly sample 50 color fundus images from the Ichallenge-AMD dataset with corresponding synthesized FFA. These paired two modalities along with the randomly augmented fundus samples are fed into the network to get feature representations, followed by reducing the feature dimension to 2 by t-SNE~\cite{maaten2008visualizing}.}
			The closer the fundus (red circle) and the augmented image (orange circle) embedding, the better transformation-invariant features are learned.	
			The closer the fundus (red circle) and the corresponding FFA (green rectangle) embedding, the better modality-invariant features are learned.	
			The detailed results for each variant are listed in Table~\ref{tab:ablation_study}.   
			Best viewed in color.}\label{fig:feature} 
		
	\end{figure*}

	\subsection{Comparison on Generalization Capacity}
	To demonstrate the generalizable features, we show the transfer learning results of our method. 
	We pre-train the self-supervised model on the EyePACS dataset and fine-tune the model on the Ichallenge-AMD and Ichallenge-PM dataset, respectively. In the pre-training stage, we do not utilize any labels while the labels are required in the fine-tuning stage.
	Our goal is to investigate whether our self-supervised method can learn more generalizable or transferable features that can be easily transferred to other tasks.  
	We compared with the state-of-the-art self-supervised method~\cite{ye2019unsupervised}. 
	During the pre-training stage, we trained all the self-supervised methods until convergences (around 200 epochs) on the EyePACS dataset. 
	Then, the learned network weight is employed as the network initialization and is fine-tuned on the Ichallenge-AMD and Ichallenge-PM dataset, respectively. 
	During the fine-tuning stage, all the models are trained with the same learning strategy and data augmentation, and the only difference is the network initialization.

	Table~\ref{tab:transfer} lists the transfer learning results on the Ichallenge-AMD and Ichallenge-PM dataset. \emph{Random init} denotes the network is trained with randomly weight initialization. 
	From Table~\ref{tab:transfer} we can see that our method consistently outperforms the state-of-the-art self-supervised method~\cite{ye2019unsupervised} on two benchmark datasets. As for the Ichallenge-AMD dataset, it is observed that our method can achieve around 6.0\% and 1.5\% improvement on AUC over \emph{Random init} and \emph{Invariant}~\cite{ye2019unsupervised}, respectively. 
	Similarly, our result on the Ichallenge-PM dataset also excels \emph{Random init} and \emph{Invariant}. 
	These consistent results demonstrate the excellent transfer learning capacity of our method.

	\subsection{Analytical Studies}
	
	\subsubsection{Comparison to other alternatives}
	To show the effectiveness of our method, we compare the following variants: \textbf{Enlarged-Data}: \revise{Train a self-supervised model with the method \cite{ye2019unsupervised}, where the multi-modal data is used by simply enlarging the dataset. For example, the enlarged dataset has 2$n$ samples, where $n$ is original color fundus and $n$ is corresponding FFA. All 2$n$ samples are used as the training images. }   
	\textbf{As-Augmentation}: Train a self-supervised model on the instance discrimination task~\cite{ye2019unsupervised} by adding the synthesized modality data as an augmentation.
	\textbf{Ours}: Train a self-supervised model on multi-modal data with constraint in Eq.~(\ref{eq:loss_total}). 
	
	Figure~\ref{fig:feature} visualizes the learned feature embedding of three variants.
	We randomly sample 50 color fundus images from the Ichallenge-AMD dataset with corresponding synthesized FFA. These images along with the randomly augmented samples are fed into the network to get feature representations, followed by reducing the feature dimension to 2 by t-SNE~\cite{maaten2008visualizing}.
	The closer the fundus and the augmented fundus image embedding, the better transformation-invariant feature is learned.
	Similarly, the closer the fundus and FFA image embedding, the better modality-invariant feature is learned. 
	We can see from Figure~\ref{fig:feature}(a) that \emph{Enlarged-Data} achieves inferior performance and there is no apparent relationship between color fundus and FFA images. 
	This is because both the fundus image and FFA learn the transformation-invariant features on its own, and the cross-modality information is neglected by the network.   
	\emph{As-Augmentation} can close the distance between the color fundus and the corresponding FFA images (the red circle and the green rectangle in Figure~\ref{fig:feature}(b)), but the performance is still inferior. 
	It is observed from Figure~\ref{fig:feature}(c) that our method can minimize the distance among fundus, transformed image, and FFA image, and at the same time enlarge the distance among different patients.
	Results for each variant on the Ichallenge-AMD dataset are summarized in Table~\ref{tab:ablation_study}. We can see our method can outperform \emph{Enlarged-Data} and \emph{As-Augmentation} on all five metrics. In particular, our result surpasses \emph{Enlarged-Data} and \emph{As-Augmentation} by around $8.86\%$ and $3.6\%$ on AUC, respectively. 
	These comparisons show that the modality-invariant constraint on multi-modal data is very useful and can contribute to better feature representation.
	
	\begin{table}[t]
		\centering
		\caption{Ablation study on the Ichallenge-AMD dataset.
			(a) Enlarged-Data: Train a self-supervised model with multi-modal data by simply enlarging the dataset.   
			(b) As-Augmentation: Train a self-supervised model on the instance discrimination task by adding the synthesized modality data as an augmentation.
			(c) Ours: Train a self-supervised model on multi-modal data with constraint in Eq.~(\ref{eq:loss_total}). 
			(Unit: \%)}
		\resizebox{0.5\textwidth}{!}
		{\begin{tabular}{c|c|c|c|c|c}
				\toprule[1.5pt]
				& AUC & Accuracy & Precision & Recall & F1-score 
				\tabularnewline \hline 	
				(a) Enlarged-Data & 65.72 & 80.93 & 70.97 & 65.72 & 67.35
				\tabularnewline 
				(b) As-Augmentation & 70.93 & 83.21 & 77.78 &  70.93 & 72.65 
				\tabularnewline 
				(c) \textbf{Ours} & \textbf{74.58} &  \textbf{86.58} & \textbf{83.20} & \textbf{74.58} &	\textbf{77.33}		
				\tabularnewline 
				\bottomrule[1.5pt]
			\end{tabular}
			\label{tab:ablation_study}}
	\end{table}
	
	\begin{table}[t]
		\centering
		\caption{\revise{Analysis of supervised learning on the Ichallenge-AMD dataset. ``Fundus'' denotes the color fundus and ``Syn FFA'' denotes the synthesized FFA. (Unit: \%).} }
		\resizebox{0.5\textwidth}{!}
		{\begin{tabular}{c|c|ccccc}
				\toprule[1.5pt]
				Modality & Backbone	&  AUC & Accuracy & Precision & Recall & F1-score 
				\tabularnewline \hline 				
				Fundus	 & 	\multirow{2}{*}{ResNet18}	 & 77.19 & 87.09 & 82.98 & 77.82 & 79.27
				\tabularnewline  
				Syn FFA &   &   77.05 & 83.97 & 76.44 &  77.05 & 76.28
				\tabularnewline \hline 
				%
				

				Fundus & \multirow{2}{*}{ResNet34} & 77.79 & 86.83 & 82.78 & 77.79 & 79.06
				\tabularnewline  
				Syn FFA	&  &  77.35 & 84.30  & 77.48 & 77.35 & 77.06   
				\tabularnewline	\hline 
				
				Fundus & \multirow{2}{*}{ResNet50} & 75.21& 84.89& 78.80 & 75.21 & 76.00
				\tabularnewline  
				Syn FFA	&  & 75.69 & 84.98 & 78.94 & 75.69  &  76.33
				\tabularnewline	
				\bottomrule[1.5pt]
			\end{tabular}
			\label{tab:ffa}
		}
	\end{table}
	
	%
	%
	%
	%
	
	\begin{figure}[!t]
		\centering
		\includegraphics[width=0.5\textwidth]{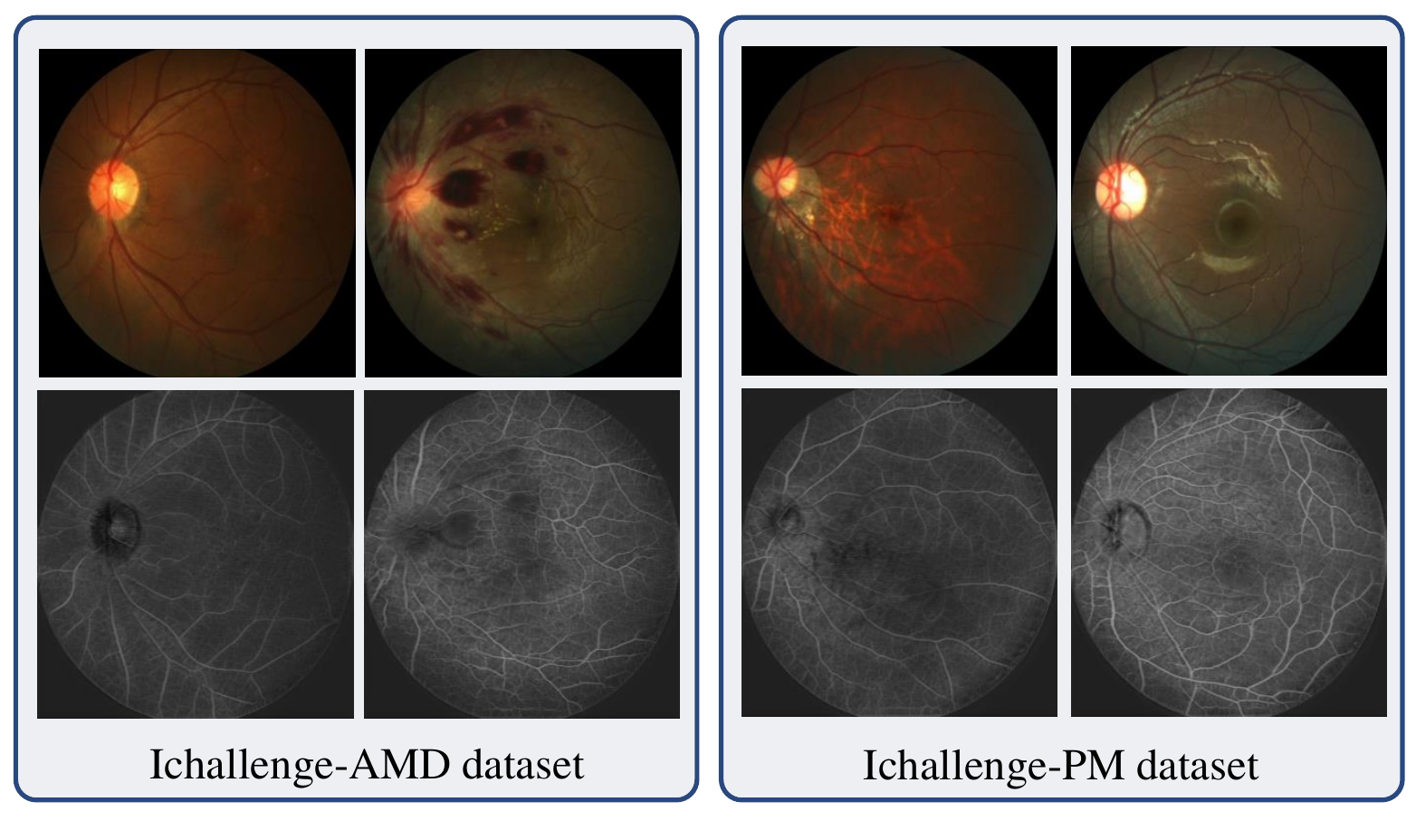}
		\caption{Visualization of synthesized FFA on the Ichallenge-AMD and Ichallenge-PM dataset. The first row denotes the color fundus images and the second row is the synthesized FFA images through our trained CycleGAN model. } 
		\label{fig:result_classify}
		
	\end{figure}
	
	\begin{figure*}[!t]
		\centering
		\includegraphics[width=0.8\textwidth]{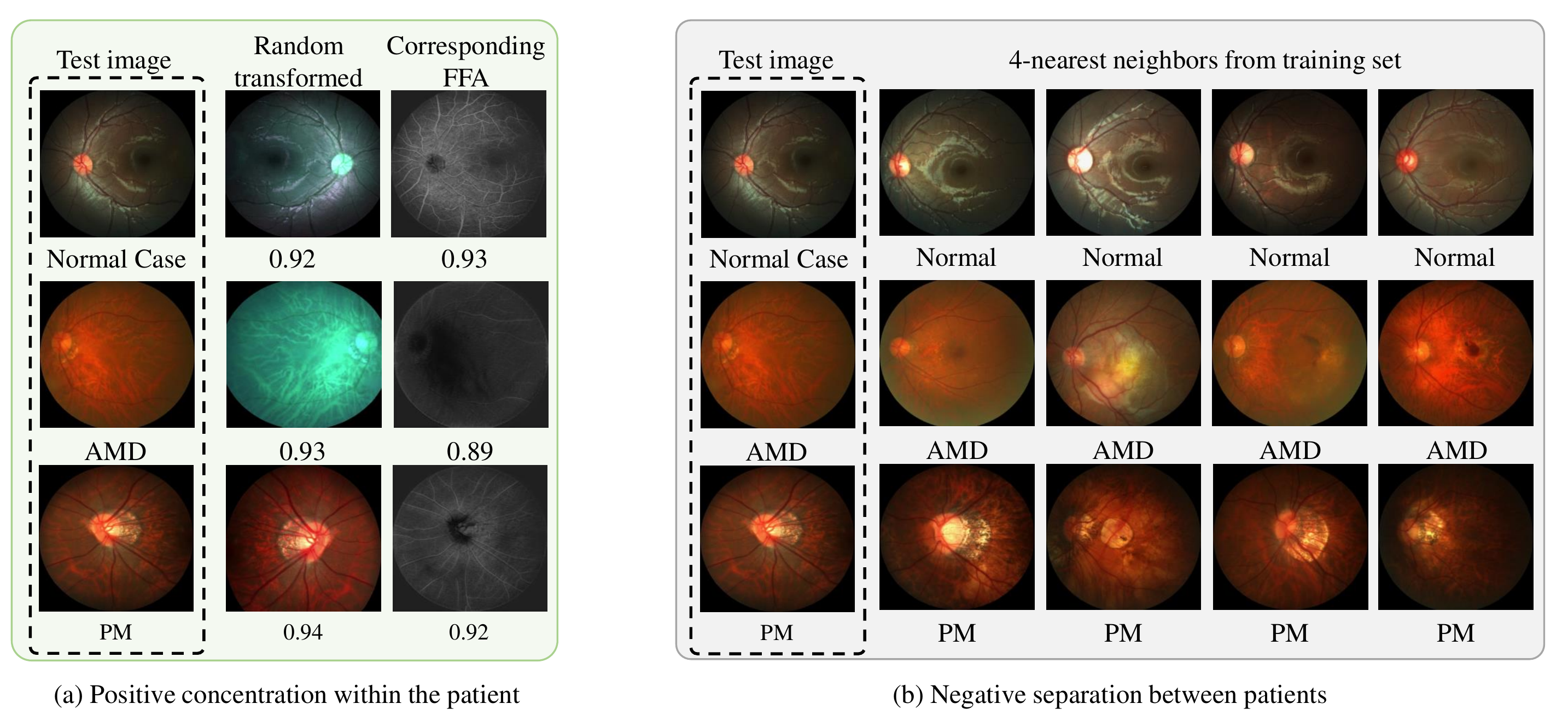}
		\caption{
			(a) Given the test images with the normal case, AMD, and PM, we first perform the random transformation to obtain the \revise{transferred samples}, and use the GAN to generate the corresponding FFA images for each test image. Note that both the random transformed images and the corresponding FFA images have high similarity scores with the test sample (numbers below each figure), which indicates the positive concentration in Eq.~(\ref{eq:1}) and (\ref{eq:syn}) can learn the \emph{transformation-} and \emph{modality-invariant} features. 
			(b) 
			We retrieve 4-nearest neighbors from training set for each test images based on the similarity scores through KNN algorithm.
			The retrieved images have high visual similarity with the test sample,  identifying the \emph{negative separation} in Eq.~(\ref{eq:3}) can capture \emph{visual similarity among patients}.} 
		\label{fig:knn}
	\end{figure*}

	\subsubsection{Visualization of patient-similarity and modality-invariant features}
	Figure~\ref{fig:feature} shows the learned feature embedding and we found that \emph{color fundus} is very close to \emph{FFA image} in the embedding space, demonstrating the learned modality-invariant features.  
	We also show the similarity score in Figure~\ref{fig:knn}(a), where the maximum similarity score is 1.0.
	We can observe that FFA achieves a high similarity score with the test image in the embedding space, which further demonstrated the modality-invariant features. 
	
	In Section I, we argue that learning to separate patients can learn apparent visual similarity among patients, \ie, patient-similarity features. To demonstrate the patient-similarity features, we visualize the K-Nearest Neighbors. In Figure~\ref{fig:knn}(b), for one test image, we visualize 4-nearest neighbors with a ground-truth label shown below each figure. We found that these neighboring images are very similar to the test image, demonstrating that our method can capture the similarity among patients, \ie, patient-similarity features.

	\subsubsection{Evaluation on the synthesis quality}
	
	Since there are no ground-truth FFA images presented in both Ichallenge-AMD and Ichallenge-PM datasets, we evaluated the quality of FFA images by running \revise{conventional supervised learning experiments with synthesized FFA images.}
	Specifically, we train the supervised baseline with the synthesized FFA images and all the training strategies are the same with those trained for color fundus images. We showed the highest supervised learning results in Table~\ref{tab:ffa}. 
	\revise{It can be observed that our synthesized FFA images can achieve similar classification results, compared to the color fundus images under three different backbones, including ResNet18, ResNet34, and ResNet50. The results demonstrate that the synthesized FFA images can obtain satisfactory quality to perform disease classification tasks. }
	We also visualize the synthesized FFA images in Figure~\ref{fig:result_classify}. We can see that the retinal vasculature can be observed in the FFA images. 
	
	\revise{
		To analyze the effects of the synthesized FFA images on our method, we run our method with different synthesized images generated at different models (saved at 450 and 500 epochs respectively). The results are shown in~Table~\ref{tab:synthesis}. 
		It is observed that the disease classification results are very similar, which indicates that our method is robust to the synthesized images generated by the trained CycleGAN network.}
	
	\begin{table}[t]
		\centering
		\caption{\revise{The effects of the synthesized images on our method. ``Syn1'' and ``Syn2'' denote that our method is trained with synthesized FFA generated at epoch 450 and 500, respectively.  (Unit: \%).} }
		
		{\begin{tabular}{c|ccccc}
				\toprule[1.5pt]
				&  AUC & Accuracy & Precision & Recall & F1-score 
				\tabularnewline \hline 				
				Syn1 & 74.45 & 85.06 & 79.59 & 74.45 & 76.31
				\tabularnewline  
				Syn2 & 74.58 &  86.58 & 83.20 & 74.58 & 77.33	
				\tabularnewline  
				\bottomrule[1.5pt]
			\end{tabular}
			\label{tab:synthesis}}
	\end{table}

	\revise{
		\subsubsection{Ablation study on our method}
		\begin{table*}[!t]
			\centering
			\caption{\revise{Ablation Study on our method. The first row denotes that the input is a doublet, consisting of ``color fundus and transformed fundus''. The second row denotes that the input is a doublet, consisting of ``color fundus and corresponding FFA''. The second row denotes that the input is a triplet, consisting of ``color fundus, transformed fundus and corresponding FFA'' (Unit: \%)}}
			
			{		
				\begin{tabular}{cc|c|ccccc}
					\toprule[1.5pt]
					
					\multicolumn{2}{c|}{Positive} & Negative & \multirow{2}{*}{AUC} & \multirow{2}{*}{Accuracy} & \multirow{2}{*}{Precision} & \multirow{2}{*}{Recall} &  \multirow{2}{*}{F1-score}
					\tabularnewline \cline{1-3} 
					Transformation-invariant & Modality-invariant & Patient-similarity & &&&& 
					\tabularnewline  	\hline 
					\checkmark & & \checkmark &	71.42  & 84.31  & 77.99  & 71.42  & 73.67
					\tabularnewline  
					& \checkmark & \checkmark & 70.48	& 83.80 & 76.90 & 70.48 & 72.68 
					
					\tabularnewline 
					\checkmark & \checkmark & \checkmark &  \textbf{74.58} &  \textbf{86.58} & \textbf{83.20} & \textbf{74.58} & 	\textbf{77.33}		
					\tabularnewline
					
					\bottomrule[1.5pt]
				\end{tabular}
				\label{tab:result_ablation}}
		\end{table*}
		Our method is a contrastive loss function performed on the triplets to optimize ``positive pairs'' and ``negative pairs''. This function optimizes the joint probability of positive pairs and negative pairs; see Eq.(4). ``positive pairs'' are the same as the ``correct prediction'' that should be minimized while ``negative pairs'' are similar to the ``wrong prediction'' that should be maximized. Hence, the optimization must be done with at least one positive and one negative pair.
		
		Since our method has two positive pairs and one negative pair, we conduct ablation study on different combinations of positive and negative pairs, and results are shown in Table~\ref{tab:result_ablation}. The positive pairs denote that the features should be pulled together, where transformation-invariant and modality-invariant features can be learned. The negative pairs denote that the features should be separated away, where patient-similarity features can be learned. It is observed that our method achieves the best performance when learning both transformation-invariant and modality-invariant features.}

	\subsubsection{Analysis on modality-specific features}
	In Eq~(\ref{eq:syn}), we encourage the network to learn modality-invariant features. 
	However, it is widely known that different modalities present modality-specific information. 
	In this section, we investigate the effectiveness of preserving modality-specific information for unsupervised representation learning.
	We implemented two variants to persevere the modality-specific information. 
	In the first variant, we followed multi-domain self-supervised learning work~\cite{feng2019self} and implemented an auxiliary classification branch to differentiate the modalities. We namely this experiment as \emph{multi-task} and the result is shown in Table~\ref{tab:specific}. 
	However, we found that this multi-task approach would hurt the performance.
	In the second variant, we added a margin in the numerator in Eq.~(\ref{eq:syn}) to control the concentration of modality-invariant features. A large margin denotes less concentration on positive pairs. 
	However, as results in Table~\ref{tab:specific}, we found there is no apparent performance improvement by learning modality-specific features.

	\begin{table}[t]
		\centering
		\caption{Ablation study on modality-specific features. 
			(Unit: \%)}
		{\begin{tabular}{c|c|c|c|c|c}
				\toprule[1.5pt]
				& AUC & Accuracy & Precision & Recall & F1-score 
				\tabularnewline \hline 	
				multi-task~\cite{feng2019self} & 67.20 & 79.66 & 70.08 & 67.20 & 68.06 
				\tabularnewline 
				
				margin0.2 & 73.54 & 83.88 & 78.14 & 73.54 & 74.71
				\tabularnewline 
				
				margin0.1 & 74.82 & 84.81 & 78.65  & 74.82 & 75.97
				\tabularnewline 
				\textbf{Ours}  & \textbf{74.58} &  \textbf{86.58} & \textbf{83.20} & \textbf{74.58} &	\textbf{77.33}		
				\tabularnewline 
				\bottomrule[1.5pt]
			\end{tabular}
			\label{tab:specific}}
	\end{table}
	
	\revise{\subsubsection{Statistic analysis}
		To provide the statistic analysis on our method, we perform the independent t-tests on Invariant~\cite{ye2019unsupervised} and our method on the Ichallenge-AMD dataset. 
		We run each experiment three times with randomly initialized seed. Through the t-test, $p$-value is 0.00064, which is significantly smaller than 0.05.
		The result indicates strong evidence that there is more than a 95\% probability that our method is statistically better than Invariant~\cite{ye2019unsupervised}. 
	}

\section{Discussion}

Recently, with the advances of deep learning techniques, automatic retinal disease diagnosis have been well studied 
in the research community, such as AMD classification~\cite{burlina2016detection,burlina2017automated,grassmann2018deep,peng2019deepseenet}, DR grading~\cite{li2019canet,sakaguchi2019fundus,wang2017zoom,zhou2019collaborative} and PM classification~\cite{2020paml}, etc.
Although satisfactory results were achieved on these tasks, these methods require a large amount of labeled data which are difficult and expensive to obtain. 
In this work, we propose a self-supervised method for retinal disease diagnosis via effectively exploiting multi-modal data. We formulate a patient feature-based softmax embedding learning objective, where modality-invariant features and patient-similarity features are learned. Our method is validated on two public retinal disease datasets, \ie, Ichallenge-AMD and Ichallenge-PM challenge, in which our method consistently outperforms other self-supervised methods and is comparable with the supervised baseline. 
Our method also surpasses other methods in terms of transfer learning, showing the effectiveness of our method in learning generalizable and transferable features.

Although our method achieves excellent performance, it comes with limitations. Since the number of fundus-FFA images is limited, we compromise to develop a multi-modal self-supervised model by synthesizing the FFA images. 
In reality, FFA images would provide more information about microaneurysms and hemorrhage, which would be beneficial for the disease diagnosis, such as AMD and PM~\cite{wang2017fundus,patel2015color,hoang2020imaging}.
One solution is to collect the datasets with color fundus and corresponding FFA images. 
Another limitation of our method is that in this paper we focus on unsupervised feature learning. 
We follow the standard evaluation protocol in most self-supervised and unsupervised learning  works~\cite{he2019momentum,henaff2019data,tian2019contrastive}, where the feature learning stage is unsupervised and the label information is required in the final classifiers, such as KNN or fully connected layer.
To make the whole diagnosis process unsupervised, one solution is to investigate joint learning of feature embedding and estimation of cluster assignments (or
labels).
In particular, we will consider to connect the feature learning with the soft and regularized deep K-means algorithm~\cite{jabi2019deep}.


The future direction we would like to work on is to better
model the mutual information between multi-modal data. 
Another potential research direction is to extend our method to more multi-modal medical imaging applications, such as multi-modal MRI, CT-MRI recognition tasks, etc.  
Even though only one modality is available in some cases, we can synthesize another modality through adversarial learning. 
Through this, we hope to leverage the general feature representation to improve a lot of downstream tasks, such as segmentation, classification, and detection~\cite{zhou2019models,li2018h}. Also, it might bring some new insights to computer-aided diagnosis in an unsupervised way.

\section{Conclusion}

This paper presents a novel self-supervised learning method by effectively exploiting multi-modal data for disease diagnosis from fundus images. 
Our key idea is to jointly utilize two modalities,~\ie, color fundus, and FFA, to learn better feature representation. Our proposed patient feature-based softmax embedding can achieve this goal by learning modality-invariant features and patient-similarity features, which show effective for fundus disease classification. Experimental results on two public datasets demonstrate that our method outperforms other self-supervised methods and achieves comparable performance to the supervised baseline. 
We also show the excellent performance of our method in learning generalizable features. 

\bibliographystyle{IEEEtranN}
\small{\bibliography{refs}}
\ifCLASSOPTIONcaptionsoff
  \newpage
\fi
\end{document}